\documentclass[10pt,journal,compsoc]{IEEEtran}
\ifCLASSOPTIONcompsoc
  \usepackage[nocompress]{cite}
\else
  \usepackage{cite}
\fi
\usepackage{hyperref} 
\hypersetup{
    colorlinks=true,
    linkcolor=red,
}
\usepackage{amssymb}
\usepackage{bbding}
\usepackage{amsmath}
\usepackage{booktabs}

\ifCLASSINFOpdf
   \usepackage[pdftex]{graphicx}
\else
\fi

\usepackage{stfloats}
\usepackage{color}
\usepackage[dvipsnames]{xcolor}

\usepackage{multirow}

\usepackage{caption}
\usepackage{subcaption}
\usepackage{amsmath}
\usepackage{cprotect}

\usepackage{amssymb}
\usepackage{pifont}

\begin{document}
%
\title{
A Survey on Evaluation of Multimodal Large Language Models
}

\author{
Jiaxing~Huang and Jingyi~Zhang
\IEEEcompsocitemizethanks{\IEEEcompsocthanksitem All authors are with the College of Computing and Data Science, Nanyang Technological University, Singapore.\protect
}
}

%
%

\markboth{Journal of \LaTeX\ Class Files,
August~2024}%
{Shell \MakeLowercase{\textit{et al.}}: Bare Demo of IEEEtran.cls for Computer Society Journals}
%



\IEEEtitleabstractindextext{%
\begin{abstract} 
Multimodal Large Language Models (MLLMs) mimic human perception and reasoning system by integrating powerful Large Language Models (LLMs) with various modality encoders (e.g., vision, audio), positioning LLMs as the "brain" and various modality encoders as sensory organs. This framework endows MLLMs with human-like capabilities, and suggests a potential pathway towards achieving artificial general intelligence (AGI). 
With the emergence of all-round MLLMs like GPT-4V and Gemini, a multitude of evaluation methods have been developed to assess their capabilities across different dimensions.
This paper presents a systematic and comprehensive review of MLLM evaluation methods, covering the following key aspects: 
(1) the background of MLLMs and their evaluation; 
(2) ``what to evaluate" that reviews and categorizes existing MLLM evaluation tasks based on the capabilities assessed, including general multimodal recognition, perception, reasoning and trustworthiness, and domain-specific applications such as socioeconomic, natural sciences and engineering, medical usage, AI agent, remote sensing, video and audio processing, 3D point cloud analysis, and others;
(3) ``where to evaluate" that summarizes MLLM evaluation benchmarks into general and specific benchmarks;
(4) ``how to evaluate" that reviews and illustrates MLLM evaluation steps and metrics;
Our overarching goal is to provide valuable insights for researchers in the field of MLLM evaluation, thereby facilitating the development of more capable and reliable MLLMs. 
We emphasize that evaluation should be regarded as a critical discipline, essential for advancing the field of MLLMs. 

\end{abstract}

}

\maketitle

\IEEEdisplaynontitleabstractindextext

%
\IEEEpeerreviewmaketitle

\IEEEraisesectionheading{\section{Introduction}\label{sec:introduction}}

Artificial Intelligence (AI) has long been a challenging area of research in computer science, with the goal of enabling machines to perceive, comprehend, and reason like humans. In recent years, Large Language Models (LLMs) have made significant advancements in AI, achieving notable success across various tasks. By scaling up both data and model size, LLMs have exhibited extraordinary emergent abilities, such as instruction following, in-context learning, and chain-of-thought reasoning. Despite their superior performance on numerous natural language processing tasks, LLMs are inherently limited to the language modality, which restricts their ability to understand and reason beyond discrete text.

On the other hand, humans sense the world via multiple channels, such as vision and language, each of which has a unique advantage in representing and communicating specific concepts. This multimodal perception manner facilitates a comprehensive understanding of the world and suggests a potential pathway toward artificial general intelligence (AGI). To bridge the gap between human perception and artificial intelligence, Multimodal Large Language Models (MLLMs) have been developed to mimic human multimodal sensing capabilities.
Specifically, MLLMs position powerful Large Language Models (LLMs) as the brain, with various modality encoders serving as sensory organs, where the modality encoders enable MLLM to perceive and understand the world through multiple modalities, while the LLMs provide advanced reasoning capabilities over the complex and comprehensive multimodal information. 
This design allows MLLMs to learn to sense and reason like humans, leveraging information from multiple channels (e.g., vision, language, audio, etc.) to achieve exceptional proficiency in multimodal understanding and reasoning. As a result, MLLMs demonstrate versatile capabilities in both traditional visual tasks and more complex multimodal challenges.

As we progress toward AGI-level MLLMs, evaluation plays a crucial role in their research, development, and deployment.
Firstly, a well-designed evaluation framework can provide a more accurate reflection of MLLM capabilities, allowing for the quantification of their strengths and limitations. For instance, \cite{li2023seed} shows that while current MLLMs excel at global image comprehension, they perform less effectively in reasoning about local image regions. Similarly, \cite{nie2024mmrel} indicates that existing MLLMs struggle with fine-grained visual relation and interaction understanding.
Second, evaluating MLLMs from the perspective of trustworthiness is essential to ensuring robustness and safety, particularly in sensitive applications like medical diagnostics and autonomous driving, where reliability is paramount.
Third, exploring and evaluating MLLMs across various downstream tasks aids in their application and deployment, ensuring that they meet the specific demands of different use cases.

In summary, more comprehensive and systematic evaluation methods are essential for inspiring the development of more powerful and robust MLLMs. As MLLMs become more advanced, they, in turn, necessitate high-standard, comprehensive evaluation benchmarks. This reciprocal relationship between the evolution of MLLMs and their evaluation processes resembles a double helix, where each advances the other.
Following pioneering MLLMs like GPT-4V, BLIP, Gemini and LLava, numerous evaluation protocols have been introduced, which focus on a wide range of aspects, from assessing general multimodal capabilities in recognition, perception, and reasoning, to evaluating specific abilities in downstream applications such as socioeconomics, natural sciences and engineering, medical usage, remote sensing, etc.

Despite the significant value and interest in MLLM evaluation for supporting MLLM research, development, and deployment, the community is short of a systematic survey that can offer a big picture about current MLLM evaluation methods, existing challenges, and potential future directions. 
This paper aims to fill this gap by conducting a thorough survey of MLLM evaluation methods across a diverse range of tasks, which are categorized based on the model capabilities being examined, including general capabilities in multimodal understanding and trustworthiness, as well as specific capabilities in downstream applications such as socioeconomics, natural sciences and engineering, medical usage, remote sensing, video, audio, and 3D point cloud analysis, among others.
We conduct this survey from different perspectives, ranging from the background of MLLMs and their evaluation, to what to evaluate, where to evaluate, how to evaluate, comparative analysis, and current challenges and open directions.
We hope this survey will provide the community with a comprehensive overview on what has been accomplished, what are current challenges, and what are promising directions for MLLMs and their evaluation.

We summarize the main contributions of this work in three key aspects.
\textit{First}, we provide a systematic and comprehensive review of multimodal large language model evaluation by developing a taxonomy of existing evaluation methods and highlighting their major contributions, strengths, and limitations. This taxonomy categorizes evaluation methods based on their examined capabilities and target applications. Unlike previous surveys focused on NLP~\cite{minaee2024large, hadi2023survey} or MLLM design~\cite{zhang2024vision}, our work uniquely centers on the evaluation of MLLMs, which, to the best of our knowledge, has not been comprehensively reviewed.
\textit{Second}, we investigate and analyze the latest advancements in MLLMs and their evaluation by conducting a thorough benchmarking and discussion of existing MLLMs across multiple datasets.
\textit{Third}, we identify and discuss several challenges and promising directions for future research in both MLLMs and their evaluation.

\begin{figure*}[t]
\centering
\includegraphics[width=0.9\textwidth]{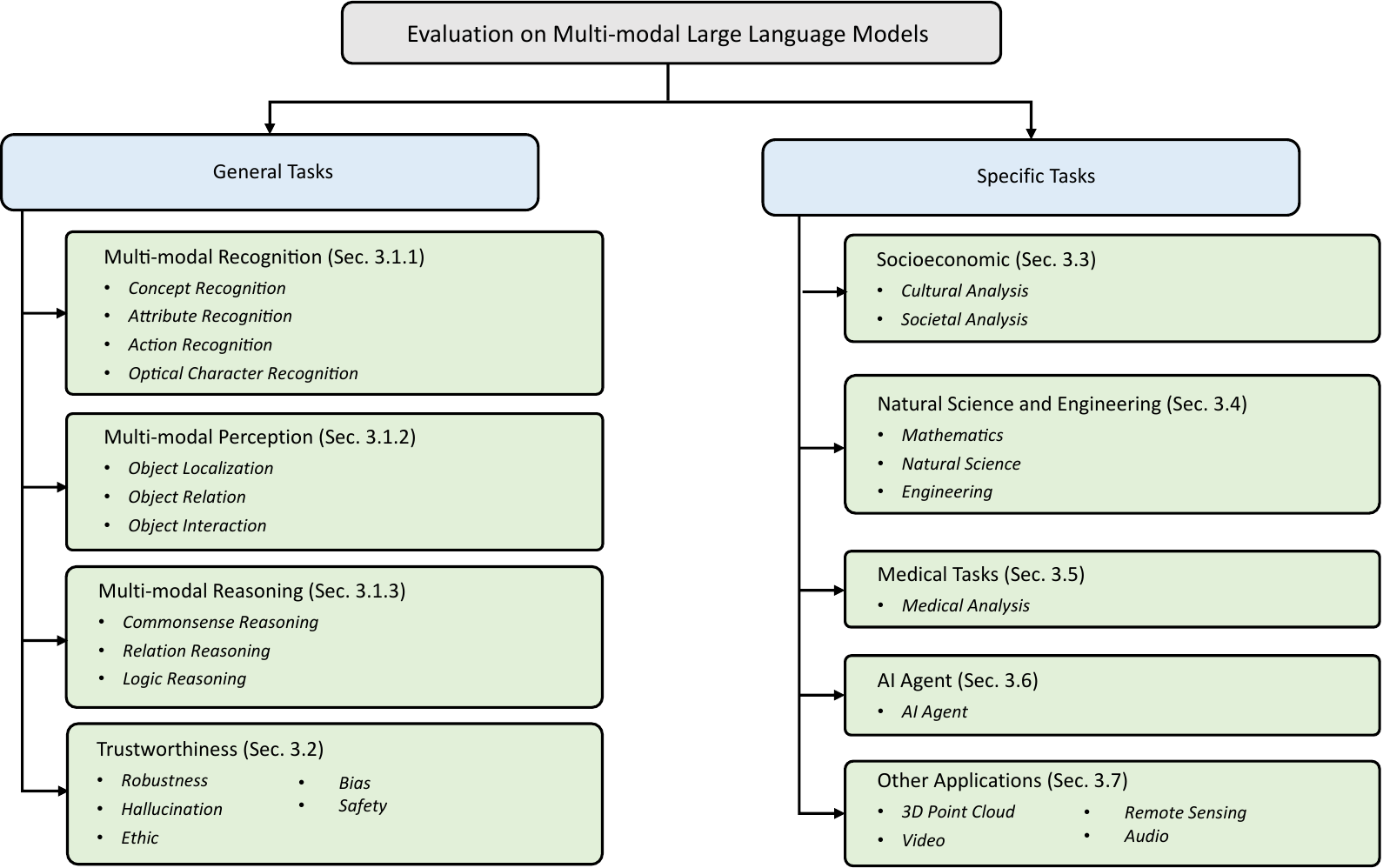}
    \caption{Typology of evaluation on multimodal large language models.
    }
    \label{fig.intro}
\end{figure*}

\section{Background}\label{Background}

This section introduces the background of the evaluation of multimodal large language models, including the foundation of multimodal large language models and xxx.

\subsection{Multimodal Large Language Model}

This section introduces the foundation of multimodal large language models(MLLMs) that involves MLLM frameworks and MLLM training strategy, and the evaluation on MLLM.

\subsubsection{MLLM Framework}

MLLMs typically consist of a large-language model that processes input texts, a modality encoder that encodes the inputs of other various modalities (e.g., image, video), and a modality projector that aligns text inputs and inputs of other modalities into a aligned feature space.

\noindent\textbf{Large Language Models.} 
For learning the input texts, transformer-based large language models (LLMs) are widely adopted. Specifically, the Transformer architecture~\cite{vaswani2017attention} employs an encoder-decoder framework, where the encoder consists of six layers, each featuring a multi-head self-attention mechanism and a multi-layer perceptron (MLP). The decoder adopts similar structure, with six layers that incorporate multi-head attention, masked multi-head attention, and an MLP.
Building on this foundation, LLaMA~\cite{touvron2023llama} has become a leading model for text feature extraction due to its strong performance across diverse language tasks.
Further extending the LLaMA architecture, instruction-tuned models like Vicuna~\cite{zheng2023judging} and Guanaco~\cite{dettmers2023qlora} have been developed and are utilized for extracting text features in constructing MLLMs.

\noindent\textbf{Modality Encoders.} 
Various encoders are employed for processing the inputs of different modalities, such as image, video and audio.
The Vision Transformer (ViT) is widely employed for image feature extraction, utilizing a series of Transformer blocks, each composed of a multi-head self-attention layer and a feed-forward network.
In practice, various pre-trained versions of ViT are adopted based on specific application needs. For example, CLIP-pre-trained ViT is commonly used for general image understanding~\cite{liu2023visual}, while SAM-pre-trained ViT is preferred for detailed and fine-grained image analysis~\cite{zhao2023bubogpt}.

For video data encoding, ViT is enhanced with temporal encoders to capture time-related information effectively. For instance, Valley~\cite{luo2023valley} incorporates a temporal modeling component to better understand the dynamic aspects of video inputs.
For 3D image feature extraction, especially in PointCloud data, specialized models such as Point-BERT~\cite{yu2021point} and PointNet~\cite{qi2017pointnet} are used. These models are specifically designed to efficiently capture features from 3D data, enabling a more comprehensive understanding of spatial structures.

Transformer-based architectures have also been widely adopted for audio data encoding. For instance, the Whisper model~\cite{radford2023robust}, designed for general-purpose speech recognition, leverages transformer networks to learn audio features effectively.

\noindent\textbf{Modality Projector.}
In multi-modal large language models, a modality projector is commonly used to align features from different modalities (e.g., text, image, audio) into a aligned feature space. This module typically involves linear layers or neural networks designed to transform the high-dimensional input features into a unified representation. 
For example, LLaVA~\cite{liu2023visual} employs a trainable projection matrix to convert encoded visual feature into the language embedding tokens space.
By projecting each modality into a common space, the model can better capture cross-modal relationships, ensuring compatibility and alignment across diverse modalities.

\subsubsection{MLLM Training Strategy}

\textbf{Alignment Pre-training.}
As the first stage of MLLM training, alignment pre-training typically focuses on aligning different modalities and learn multimodal correspondence knowledge. Generally, the pre-training involves large-scale text-paired data, such as captions that describe images, audio, or videos in natural language.
For example,~\cite{liu2023visual,liu2023improved} employs a standard cross-entropy loss for enabling MLLMs to autoregressively predicte captions for given images during the alignment pre-training stage.
For better preserving the original pre-trained knowledge, MLLMs often keep pre-trained models (e.g., pre-trained vision encoders or large-language models) frozen while only training a learnable projection module for alignment~\cite{liu2023visual,liu2023improved}.

\noindent\textbf{Multimodal Instruction Tuning.}
Multimodal instruction tuning fine-tunes MLLMs with language as task instructions, aiming for build a versatile model with superior interactivity and adaptability in following the users' intents.
The instruction tuning generally consists of two stages, i.e.,(1) visual instruction-following data construction and (2) visual instruction tuning. 
Visual instruction-following data typically have the format of \{\verb|Instruction|, \verb|Input|, \verb|Output|\}, where \verb|Instruction| denotes task instructions, \verb|Input| refers to input of various modalities (e.g., \verb|Input| = \{\verb|Image|\}) and \verb|Output| stands for the response regarding the given task instruction.
These datasets are often expanded from public multimodal data and enhanced using large language models~\cite{radford2021learning,schuhmann2021laion}.
With the constructed instruction-following data, the MLLMs are fine-tuned in a full-supervised manner by predicting each token in the output sequentially based on the instruction and input image.

\noindent\textbf{Alignment for human preference.}
Alignment tuning particularly aims to enhance model behavior to meet specific human expectations.
Two techniques for alignment tuning are widely-adopted, i.e., reinforcement learning with human feedback (RLHF)~\cite{sun2023aligning} and direct preference optimization (DPO)~\cite{lei2023learning}.
Specifically, RLHF involves training models using rewards based on human feedback, guiding them toward more desirable outputs. On the other hand, DPO directly optimizes the model by learning from human preferences, improving alignment in a more straightforward manner without requiring complex reward models.

\begin{table*}[!ht]
    \centering
    \caption{Summary of the general evaluation tasks.
    }
    \resizebox{0.99\linewidth}{!}{
    \begin{tabular}{l|p{6cm}|p{9cm}}
        \toprule[1pt]
        \textbf{Tasks}& Task Description & Related Benchmarks \\
        \midrule
        \multicolumn{3}{c}{\textbf{Multi-modal Recognition}}  \\
        \midrule
        Concept recognition & Recognizing visual concepts, e.g., objects, instances and scenes.& MMBench~\cite{liu2023mmbench}, MM-Vet~\cite{yu2023mm}, Seed-Bench~\cite{li2023seed}, MME~\cite{fu2023mme}, MMStar~\cite{chen2024we}, LLaVA-Bench~\cite{liu2023llava}, Open-VQA~\cite{zeng2024matters}, MDVP-Bench~\cite{lin2024draw}, P$^2$GB~\cite{chen2024plug}, EQBEN~\cite{wang2023equivariant}, MUIRBENCH~\cite{wang2024muirbench}, TouchStone~\cite{bai2023touchstone}, mPlug-Owl~\cite{ye2023mplug}, MMIU~\cite{meng2024mmiu}, LogicVista~\cite{xiao2024logicvista}, CODIS~\cite{luo2024codis}\\
        
        Attribute recognition &Recognizing visual subject's attributes e.g., style, quality, mood, quantity, material, and human's profession. &MMBench~\cite{liu2023mmbench}, MM-Vet~\cite{yu2023mm}, Seed-Bench~\cite{li2023seed}, V*Bench~\cite{wu2024v}, MMVP~\cite{tong2024eyes}, CV-Bench~\cite{tong2024cambrian}, Visual CoT~\cite{shao2024visual}, EQBEN~\cite{wang2023equivariant}, SPEC~\cite{peng2024synthesize}, VL-Checklist~\cite{zhao2023vlchecklist}, ARO~\cite{yuksekgonul2023and}, MUIRBENCH~\cite{wang2024muirbench}, COMPBENCH~\cite{kil2024compbench},  MME~\cite{fu2023mme}, Open-VQA~\cite{zeng2024matters}, TouchStone~\cite{bai2023touchstone}, ImplicitAVE~\cite{zou2024implicitave}, CFMM~\cite{li2024eyes}, VL-ICL~\cite{zong2024vl}, CLEVR~\cite{johnson2016clevr}, AesBench~\cite{huang2024aesbench}, UNIAA~\cite{zhou2024uniaa}, Q-Bench~\cite{wu2023q}, Q-Bench$^+$~\cite{zhang2024benchmark}, CODIS~\cite{luo2024codis}, VQAv2~\cite{goyal2017makingvvqamatter}, GQA~\cite{hudson2019gqanewdatasetrealworld}, MMStar~\cite{chen2024we}, CODIS~\cite{luo2024codis}
\\ 
        
        Action Recognition & Recognizing actions or activities performed by subjects. & MMBench~\cite{liu2023mmbench}, Seed-Bench~\cite{li2023seed}, Open-VQA~\cite{zeng2024matters}, Visual CoT~\cite{shao2024visual}, EQBEN~\cite{wang2023equivariant}, VL-Checklist~\cite{zhao2023vlchecklist}, MILEBENCH~\cite{song2024milebench},  \\
        
        Optical Character Recognition & Recognizing and converting text from visual inputs, such as images of documents or signs. & MMBench~\cite{liu2023mmbench}, MM-Vet~\cite{yu2023mm}, Seed-Bench~\cite{li2023seed}, MME~\cite{fu2023mme}, Open-VQA~\cite{zeng2024matters}, TouchStone~\cite{bai2023touchstone}, mPlug-Owl~\cite{ye2023mplug}, P$^2$GB~\cite{chen2024plug}, Visual CoT~\cite{shao2024visual}, VL-ICL~\cite{zong2024vl}, OCRBench~\cite{liu2023hidden}, TextVQA~\cite{singh2019towards}, TextCaps~\cite{sidorov2020textcaps}, Seed-bench-2-plus~\cite{li2024seed}, DocVQA~\cite{mathew2020document}, MPDocVQA~\cite{tito2023hierarchical}\\
        \midrule
        \multicolumn{3}{c}{\textbf{Multi-modal Perception}}  \\
        \midrule
        Object localization & Identifying the spatial position of objects in a scene.& MMBench~\cite{liu2023mmbench}, MM-Vet~\cite{yu2023mm}, Seed-Bench~\cite{li2023seed}, MME~\cite{fu2023mme}, MDVP-Bench~\cite{lin2024draw}, CODE~\cite{zang2023contextual}, MMVP~\cite{tong2024eyes}, P$^2$GB~\cite{chen2024plug}, Visual CoT~\cite{shao2024visual}, EQBEN~\cite{wang2023equivariant}, SPEC~\cite{peng2024synthesize}, VL-Checklist~\cite{zhao2023vlchecklist}, MILEBENCH~\cite{song2024milebench}, COMPBENCH~\cite{kil2024compbench}, MMIU~\cite{meng2024mmiu}, VSR~\cite{liu2023visual}, SpatialRGPT~\cite{cheng2024spatialrgpt}, CODIS~\cite{luo2024codis}, CFMM~\cite{li2024eyes}, MMStar~\cite{chen2024we}, M3GIA~\cite{song2024m3gia}, LogicVista~\cite{xiao2024logicvista}, CODIS~\cite{luo2024codis}\\
        
        Object relation & Understanding object spatial relations e.g., "before," "next to," "left of," "right of," etc.& MMBench~\cite{liu2023mmbench}, MM-Vet~\cite{yu2023mm}, Seed-Bench~\cite{li2023seed}, MME~\cite{fu2023mme}, MDVP-Bench~\cite{lin2024draw}, V*Bench~\cite{wu2024v}, MMVP~\cite{tong2024eyes}, CV-Bench~\cite{tong2024cambrian}, Visual CoT~\cite{shao2024visual}, VL-Checklist~\cite{zhao2023vlchecklist}, ARO~\cite{yuksekgonul2023and}, MMIU~\cite{meng2024mmiu}, MMRel~\cite{nie2024mmrel}, What’sUp~\cite{kamath2023s}, GSR-BENCH~\cite{rajabi2024gsr}, CRPE~\cite{wang2024all}, VSR~\cite{liu2023visual}, SpatialRGPT-Bench~\cite{cheng2024spatialrgpt}, CODIS~\cite{luo2024codis}, CLEVR~\cite{johnson2016clevr}, GQA~\cite{hudson2019gqanewdatasetrealworld}, MMStar~\cite{chen2024we}, M3GIA~\cite{song2024m3gia}, LogicVista~\cite{xiao2024logicvista}, CODIS~\cite{luo2024codis}\\
        
        Object interaction & Understanding how objects interact with each other or with agents. & Seed-Bench~\cite{li2023seed}, P$^2$GB~\cite{chen2024plug}, VL-Checklist~\cite{zhao2023vlchecklist}, ARO~\cite{yuksekgonul2023and}, CODIS~\cite{luo2024codis} \\
        \midrule
        \multicolumn{3}{c}{\textbf{Multi-modal reasoning} } \\
        \midrule
        
        Commonsense Reasoning & Applying general world knowledge to infer logical conclusions. & MMBench~\cite{liu2023mmbench}, MM-Vet~\cite{yu2023mm}, MME~\cite{fu2023mme}, LLaVA-Bench~\cite{liu2023visual}, Open-VQA~\cite{zeng2024matters}, TouchStone~\cite{bai2023touchstone}, MDVP-Bench~\cite{lin2024draw}, M$^3$CoT~\cite{chen2024m}, II-Bench~\cite{liu2024ii}, CFMM~\cite{li2024eyes}, VL-ICL~\cite{zong2024vl}, MMStar~\cite{chen2024we}, FVQA~\cite{wang2017fvqa}, OK-VQA~\cite{marino2019ok}, A-OKVQA~\cite{schwenk2022okvqa}, MIKE~\cite{li2024mike}, VLKEB~\cite{huang2024vlkeb}, MC-MKE~\cite{zhang2024mc}, M3GIA~\cite{song2024m3gia}, MMMU~\cite{yue2024mmmu}
 \\ 
        
        Relation Reasoning & Understanding complex social, physical, or natural relationships. &MMBench~\cite{liu2023mmbench}, Visual CoT~\cite{shao2024visual}, II-Bench~\cite{liu2024ii}, RAVEN~\cite{zhang2019raven}, MMMU~\cite{yue2024mmmu}\\ 
        
        Logic Reasoning & Applying structured thinking, rules, and cause-and-effect analysis to draw conclusions, make predictions, or solve problems based on sequential or relational information. &MMBench~\cite{liu2023mmbench},Seed-Bench~\cite{li2023seed}, TouchStone~\cite{bai2023touchstone}, mPlug-Owl~\cite{ye2023mplug}, MDVP-Bench~\cite{lin2024draw}, P$^2$GB~\cite{chen2024plug}, Visual CoT~\cite{shao2024visual}, M$^3$CoT~\cite{chen2024m}, Mementos~\cite{wang2024mementos}, MILEBENCH~\cite{song2024milebench}, MUIRBENCH~\cite{wang2024muirbench}, COMPBENCH~\cite{kil2024compbench}, MMIU~\cite{meng2024mmiu}, II-Bench~\cite{liu2024ii}, MMStar~\cite{chen2024we}, RAVEN~\cite{zhang2019raven}, MARVEL~\cite{jiang2024marvel}, MaRs-VQA~\cite{cao2024visual}, MMMU~\cite{yue2024mmmu}, MM-NIAH~\cite{wang2024needle}, ChartQA~\cite{masry2022chartqa}, ChartX~\cite{xia2024chartx}, ChartBench~\cite{xu2023chartbench}, SciGraphQA~\cite{li2023scigraphqa},MMC~\cite{liu2023mmc}, CHarxiv~\cite{wang2024charxiv}, LogicVista~\cite{xiao2024logicvista}\\ 
        \midrule
        \multicolumn{3}{c}{\textbf{Trustworthiness}}  \\
        \midrule

        {Robustness} & The capability of MLLMs to maintain performance under various conditions, including adversarial inputs or noisy environments. & CHEF~\cite{shi2023chef}, MAD-Bench~\cite{qian2024easy},  MMR~\cite{liu2024seeing}, MM-SpuBench~\cite{ye2024mm}, BenchLMM~\cite{cai2023benchlmm}, Multi-Trust~\cite{zhang2024benchmarking} \\

        {Hallucination} & The tendency of MLLMs to generate information that is incorrect, irrelevant, or fabricated. & POPE~\cite{li2023evaluating}, UNIHD~\cite{chen2024unified}, VideoHallucer~\cite{wang2024videohallucer},  CAP2QA~\cite{cha2024visually}, CHEF~\cite{shi2023chef}, GAVIE~\cite{liu2023mitigating}, HaELM~\cite{wang2023evaluation}, M-HalDetect~\cite{gunjal2024detecting}, Bingo~\cite{cui2023holistic}, HallusionBench~\cite{guan2024hallusionbench}, AMBER~\cite{wang2023llm}, MM-SAP~\cite{wang2024mm}, VHTest~\cite{huang2024visual}, CorrelationQA~\cite{han2024instinctive},  \\

        {Ethic} & The adherence of MLLMs to ethical guidelines, ensuring outputs align with moral and societal values. & Multi-Trust~\cite{zhang2024benchmarking}\\

        {Bias} & The presence and extent of unfair biases in the MLLM’s predictions, which could lead to discrimination or skewed results. & Multi-Trust~\cite{zhang2024benchmarking}, RTVLM~\cite{li2024red} \\

        {Safety} & The potential risks posed by the MLLM, such as generating harmful content, promoting dangerous behavior, or being misused. & MM-SafetyBench~\cite{liu2023mm}, MMUBench~\cite{li2024single}, Jailbreakv-28k~\cite{luo2024jailbreakv}, Shield~\cite{shi2024shield}, RTVLM~\cite{li2024red}, Multi-Trust~\cite{zhang2024benchmarking},\\

        \bottomrule[1pt]
    \end{tabular}
    }
    \label{tab.eval_datasets_1}
\end{table*}

\begin{table*}[!ht]
    \centering
    \caption{Summary of the specific evaluation tasks.
    }
    \resizebox{0.99\linewidth}{!}{
    \begin{tabular}{l|p{6cm}|p{9cm}}
        \toprule[1pt]
        \textbf{Tasks}& Tasks Description & Related Benchmarks \\
        \midrule
        \multicolumn{3}{c}{\textbf{Socioeconomic}}  \\
        \midrule
        Cultural Analysis &  The capability of MLLMs in understanding cultural norms, expressions, and practices across different societies. & CVQA~\cite{romero2024cvqa} \\
        Societal Analysis &The capability of MLLMs to comprehend and analyze societal issues, trends, and dynamics& VizWiz~\cite{gurari2018vizwiz}, MM-Soc~\cite{jin2024mm}, TransportationGames~\cite{zhang2024transportationgames} \\
        \midrule
        \multicolumn{3}{c}{\textbf{Natural Science and Engineering}}  \\
        \midrule
        Mathematics & The ability of MLLMs in solving mathematics problems, equation interpretation, and numerical reasoning tasks.&MM-Vet~\cite{yu2023mm}, MathVerse~\cite{zhang2024mathverse}, NPHardEval4V~\cite{fan2024nphardeval4v}, Inter-GPS~\cite{lu2021inter}, MME~\cite{fu2023mme}, TouchStone~\cite{bai2023touchstone}, M$^3$CoT~\cite{chen2024m}, MMStar~\cite{chen2024we}, M3GIA~\cite{song2024m3gia}, MathVista~\cite{lu2023mathvista},  SceMQA~\cite{liang2024scemqa}, MULTI~\cite{zhu2024multi}, LogicVista~\cite{xiao2024logicvista}, Math-V~\cite{wang2024measuring}, MathCheck~\cite{zhou2024your},\\
        
        Natural Science &The capability of MLLMs in understanding the concepts in physics, chemistry, biology, and other science subjects.&M$^3$CoT~\cite{chen2024m}, CMMMU~\cite{zhang2024cmmmu}, ScienceQA~\cite{lu2022learn}, MMMU~\cite{yue2024mmmu}, SceMQA~\cite{liang2024scemqa}, MULTI~\cite{zhu2024multi}, Peacock~\cite{alwajih2024peacock}, LaVy~\cite{tran2024lavy}, MUIRBENCH~\cite{wang2024muirbench}, MMStar~\cite{chen2024we}, M3Exam~\cite{zhang2023m3exam}, MTVQA~\cite{tang2024mtvqa}, CVQA~\cite{romero2024cvqa}, LogicVista~\cite{xiao2024logicvista}, SciFIBench~\cite{roberts2024scifibench}\\         
        Engineering & The MLLMs' ability to assist in design, technical analysis, and problem-solving within engineering disciplines.& DesignQA~\cite{doris2024designqa}, MMMU~\cite{yue2024mmmu}, Asclepius~\cite{wang2024asclepius}\\
        \midrule
        \multicolumn{3}{c}{\textbf{Medical Tasks}}    \\
        \midrule
        Medical Analysis & The capability of MLLMs in analyzing medical data and providing diagnostic insights. & MMMU~\cite{yue2024mmmu}, M3D~\cite{bai2024m3d}, GMAI-MMBench~\cite{chen2024gmai}\\
        \midrule
        \multicolumn{3}{c}{\textbf{AI Agent}}  \\
        \midrule
        AI Agent &The MLLMs' ability to autonomously perform tasks based on multi-modal inputs, and generate and follow through with plans to achieve specific goals. &Mobile-Agent~\cite{wang2024mobile}, VisualAgentBench~\cite{liu2024visualagentbench}, EgoPlan-Bench~\cite{chen2024egoplanbenchbenchmarkingmultimodallarge}, PCA-EVAL~\cite{chen2023towards}, OpenEQA~\cite{majumdar2024openeqa},  Ferret-UI~\cite{you2024ferret}, Crab~\cite{xu2024crab}\\
        \midrule
        \multicolumn{3}{c}{\textbf{Other Applications}}  \\
        \midrule
        3D Point Cloud & Interpret and process 3D spatial data for applications like robotics or autonomous driving.&ScanQA~\cite{azuma2022scanqa}, LAMM~\cite{yin2024lamm}, M3DBench~\cite{li2023m3dbench}, SpatialRGPT~\cite{cheng2024spatialrgpt}\\
        Video  &The MLLMs' ability to understand, summarize, and reason about video content. &VideoHallucer~\cite{wang2024videohallucer}, MMBench-Video~\cite{fang2024mmbench}, SOK-Bench~\cite{wang2024sok}, MVBench~\cite{li2024mvbench} \\
        Remote Sensing&Process and analyze satellite or aerial images for environmental monitoring, agriculture, and more.& HighDAN~\cite{hong2023cross}, RSGPT~\cite{hu2023rsgpt}, MDAS~\cite{hu2023mdas}\\
        Audio & The ability of MLLMs in understanding audio, like speech recognition, audio event detection, and sound classification. &AIRBench~\cite{yang2024air}, Dynamic-superb~\cite{huang2024dynamic}, MuChoMusic~\cite{weck2024muchomusic} \\
        
        \bottomrule[1pt]
    \end{tabular}
    }
    \label{tab.eval_datasets_2}
\end{table*}

\section{What to evaluation} \label{What}

This section provides an overview of the various tasks used to evaluate the capabilities of MLLMs, encompassing general tasks like multi-modal understanding and trustworthiness analysis, as well as specific tasks in areas such as socioeconomic, natural science and engineering, medical applications, AI agents, and other vision-related applications. Table~\ref{tab.eval_datasets_1} and Table~\ref{tab.eval_datasets_2} summarize the MLLM evaluation on general tasks and specific tasks respectively.

\subsection{Multi-modal understanding}

The advent of multi-modal large language models (MLLMs) has extended the capabilities of traditional language models by enabling them to process and understand information from various modalities, such as text and images. The goal of multi-modal understanding is to assess how effectively these models can integrate and interpret information across different types of input. Specifically, the multi-modal understanding task can be broadly categorized into multi-modal recognition, multi-modal perception, and multi-modal reasoning.

\subsubsection{Multi-modal Recognition} 

Multi-modal recognition aims to identify and classify specific objects, actions, and attributes across multiple modalities. This task focuses on the model’s ability to detect and recognize the various aspects, including concept recognition, attribute recognition, action recognition, and Optical Character Recognition (OCR).

\noindent\textbf{Concept recognition} focuses on the model’s ability to identify and label various entities, instances, objects, and scenes across different modalities. This task involves recognizing both general and specific concepts such as objects within an image (e.g., identifying a `car' or `dog')~\cite{liu2023mmbench,li2023seed,yu2023mm}, instances of particular categories (e.g., a specific landmark or product)~\cite{liu2023mmbench,li2023seed,yu2023mm}, and broader scenes (e.g., a `beach' or `mountain')~\cite{li2023seed}. As the key capability of MLLMs in multi-modal understanding, MLLMs generally demonstrate superior performance over concept recognition tasks. For examples,~\cite{li2023seed} shows that most MLLMs achieve relatively high performance (e.g., $>$ 40\%) on scene understanding.
In MM-Vet~\cite{yu2023mm}, LLaVA-13B (V1.3, 336px)~\cite{liu2023llava}, achieves a score of 38.1\% in concept recognition, which indicates its ability to understand and categorize visual concepts effectively. Another model, LLaMA-Adapter v2-7B~\cite{gao2023llama}, performs slightly better with a score of 38.5\%, which benefits from its large-scale tuning data.
TouchStone~\cite{bai2023touchstone} proposed a composite score termed TouchStone Score. It reflects the model's ability to perform across all evaluated tasks, including concept recognition. Qwen-VL~\cite{bai2023qwen} stands out as the top performer in concept recognition tasks within the TouchStone framework, showing superior accuracy and consistency compared to other models.
\cite{ye2023mplug} shows that mPLUG-Owl2 outperforms other models like Qwen-VL-Chat~\cite{bai2023qwen} and InstructBLIP~\cite{dai2023instructblip}. Its high CIDEr scores~\cite{vedantam2015cider} in major datasets like COCO~\cite{lin2014microsoft} and Flickr30K~\cite{young2014image} demonstrate its superior ability to accurately recognize and describe complex visual concepts, making it a leading model in this area.

\noindent\textbf{Attribute recognition} is the task of recognizing visual subject’s attributes under different modalities. It involves recognizing style, quality, emotions, quantity, material, and human’s profession. 
In MMBench~\cite{liu2023mmbench}, the performance of MLLMs on the Attribute Recognition task varies significantly. For instance, the model InternLM-XComposer2~\cite{dong2024internlm} achieved one of the highest scores with 73.0\% accuracy, demonstrating strong capabilities in this area. On the other hand, models like OpenFlamingo v2~\cite{awadalla2023openflamingo} performed poorly, with an accuracy of only 5.3\% on this task.
In the SEED-Bench~\cite{li2023seed}, the performance of MLLMs on the task related to attribute recognition is assessed under the "Instance Attributes" dimension, which is specifically designed to evaluate a model's ability to recognize and understand the attributes of an instance. Results indicates that the model InstructBLIP Vicuna~\cite{dai2023instructblip} achieved a commendable performance in the Instance Attributes, showing its strong capability in attribute recognition. 
In the MME benchmark~\cite{fu2023mme}, the performance of MLLMs on attribute recognition tasks is assessed through specific subtasks including color, material, shape, and other descriptive features of the objects present. For example, in the Color subtask, InfMLLM~\cite{zhou2023infmllmunifiedframeworkvisuallanguage} achieved a high accuracy score, demonstrating its proficiency in recognizing color attributes of objects in images. 
In the Open-VQA~\cite{zeng2024matters}, InstructBLIP~\cite{dai2023instructblip} exhibited high performance in attribute recognition. 
Results in TouchStone~\cite{bai2023touchstone} shows that Qwen-VL~\cite{bai2023qwen} emerges as the top performer in the Attribute Recognition task within the TouchStone framework, consistently delivering high accuracy in identifying detailed object attributes. mPlug-Owl~\cite{ye2023mplug} also performs strongly, while models like PandaGPT~\cite{su2023pandagpt} lag behind, especially in complex attribute recognition scenarios.

\noindent\textbf{Action Recognition} is a task that recognizing actions or activities performed by subjects under different modalities. 
In MMBench~\cite{liu2023mmbench}, the performance of MLLMs on the Action Recognition task is evaluated under the Fine-grained Perception (cross-instance) category. The task involves recognizing human actions, including pose motion, human-object interaction, and human-human interaction. Specific models and their performances are compared, with results presented in a fine-grained manner. 
According to SEED-Bench~\cite{li2023seed}, the model InstructBLIP Vicuna~\cite{dai2023instructblip} demonstrated strong performance in the "Action Recognition" dimension, outperforming other models.
In the Open-VQA~\cite{zeng2024matters}, models like InstructBLIP~\cite{dai2023instructblip} have demonstrated strong performance in Action Recognition. 
In the Visual CoT~\cite{shao2024visual}, the performance of different MLLMs on the "Action Recognition" task varies significantly. The baseline model achieved a certain level of performance across multiple datasets. However, when employing the Visual CoT (Chain of Thought) process~\cite{wei2022chain}, the performance generally improved, especially in more complex tasks that require deeper reasoning or understanding of the visual context.
By examining the performance metrics such as accuracy percentage and rank within the Action Recognition task, researchers and practitioners can gain insights into the capabilities of different MLLMs in understanding and classifying actions. This comprehensive evaluation is crucial for the advancement of MLLMs in multimodal tasks that involve temporal dynamics and sequential understanding.

\noindent\textbf{Text Recognition} refers to recognizing and converting text from visual inputs, such as images of documents or signs.
In MMBench~\cite{liu2023mmbench}, MLLM's performance on the Text Recognition task is highlighted with specific metrics and observations. The models' accuracy varied based on their architecture and size, with some models demonstrating significantly better performance due to factors like language model choice and pretraining data. For instance, open-source models like LLaVA~\cite{liu2023llava} series and InternLM-XComposer2~\cite{dong2024internlm} showed strong performance, while other models like MiniGPT struggled more on this task.
In SEED-Bench~\cite{li2023seed}, the performance of each MLLM on Text Recognition tasks is measured by its accuracy in selecting the correct option from the multiple-choice questions, which is then compared against the ground truth answer provided by human annotators. LLaVa~\cite{liu2023llava} exhibits unparalleled capabilities in the evaluation of text recognition compared to other dimensions. 
According to the MME~\cite{fu2023mme}, models like GPT-4V~\cite{openai2024gpt4technicalreport}, Skywork-MM~\cite{SkyworkMM}, and WeMM~\cite{WeMM} achieved the top scores in the OCR task. Specifically, GPT-4V~\cite{openai2024gpt4technicalreport} demonstrated a significant advantage with a score of 185, indicating its high proficiency in recognizing and transcribing text from images.
In the Open-VQA~\cite{zeng2024matters}, models like InstructBLIP~\cite{dai2023instructblip} have shown high performance in Text Recognition tasks, indicating their proficiency in recognizing and transcribing text from images.
In Visual CoT~\cite{shao2024visual}, the baseline models generally achieve moderate accuracy in OCR tasks. The use of Visual CoT (Chain of Thought) often leads to better performance in OCR tasks. This approach allows models to break down the text recognition process into more manageable steps, which can improve accuracy and understanding.
In TouchStone~\cite{bai2023touchstone}, Qwen-VL~\cite{bai2023qwen} demonstrates superior accuracy and reliability in recognizing and reading text from images.
mPlug-Owl~\cite{ye2023mplug} stands out in OCR tasks within its framework, showing superior performance compared to other models like Qwen-VL-Chat~\cite{bai2023qwen} and InstructBLIP~\cite{dai2023instructblip}. Its ability to accurately read and understand text in various forms and contexts is evidenced by high accuracy scores on datasets like TextVQA~\cite{singh2019towards}, making it a leading model for OCR-related challenges.
By examining the performance metrics such as accuracy and rank within the Text Recognition task, researchers and practitioners can evaluate the capabilities of different MLLMs in processing and interpreting textual information from visual data. This capability is essential for applications that require text recognition and interpretation, such as automated document processing or image-based information retrieval.

\subsubsection{Multi-modal Perception} 

\noindent\textbf{Object Localization} determining the position of objects in a scene. It also includes identifying counting the number of objects and determining the orientation of the object. 
In the MMBench~\cite{chen2024gmai}, MLLMs perform at a relatively moderate level on the Object Localization task. The performance varies significantly among different models. The overall accuracy in Object Localization shows room for improvement, especially when compared to other tasks within the benchmark.
MM-Vet~\cite{yu2023mm} does not have a dedicated object localization task, it assesses related capabilities through the "Spatial awareness" category, which can give an indication of how well MLMMs perform on tasks that may include object localization as part of the broader spatial awareness capability.
In the SEED-Bench~\cite{li2023seed}, the performance of MLLMs on Object localization tasks is assessed under the "Instance Location" dimension, where the model InstructBLIP~\cite{dai2023instructblip} achieved a high accuracy in the "Instance Location" dimension, indicating its strong capability in localizing instances within images. 
According to the results in MME~\cite{fu2023mme}, models like Lion and InfMLLM~\cite{zhou2023infmllmunifiedframeworkvisuallanguage} achieved high scores in the object localization subtask.
By reviewing the performance metrics such as accuracy percentage and rank within the "Instance Location" dimension, researchers and practitioners can evaluate the precision of different MLLMs in identifying the spatial context of objects within visual scenes. It is essential for understanding and improving models' spatial understanding abilities, which is a fundamental aspect of advanced multimodal AI systems.

\noindent\textbf{Object Relation} involves the model's ability to understand and identify the spatial relationships between different objects within a visual scene. This can include spatial relationships (e.g., above, next to), interactions between objects (e.g., a person holding a book), or more complex contextual connections (e.g., understanding that a chair is meant to be sat on). The task evaluates the model's capability to correctly interpret and reason about these relationships as presented in images or videos, which is crucial for tasks such as visual reasoning, scene understanding, and more complex vision-language interactions.
In MMBench~\cite{liu2023mmbench}, the performance of MLLMs on the Object Relation task shows significant variability. Specifically, the models demonstrate varying levels of success in accurately identifying relationships between objects in visual data, which could include spatial relationships, interactions, and contextual connections. The performance metrics indicate that models like GPT-4v~\cite{openai2024gpt4technicalreport} and Qwen-VL-Max~\cite{bai2023qwen} are among the top performers in this category, displaying higher accuracy in understanding and reasoning about object relations compared to other models.
MM-Vet~\cite{yu2023mm} assesses the performance of LMMs on Object relation tasks through the "Spatial awareness" capability, using an LLM-based scoring system that provides a comprehensive metric for evaluating the accuracy and response quality of models in understanding and describing object relationships within visual scenes, where MM-ReAct-GPT-4~\cite{yang2023mm} achieves a high score in the "Spatial awareness" category, indicating its strong performance in tasks that require understanding spatial relationships.
According to the SEED-Bench~\cite{li2023seed}, models such as InstructBLIP Vicuna~\cite{dai2023instructblip} and BLIP2~\cite{li2023blip} have demonstrated strong performance in the "Spatial Relation" dimension, indicating their proficiency in understanding spatial relationships between objects.
Results of MME~\cite{fu2023mme} show that certain models have demonstrated strong performance in object relation tasks. For instance, models like WeMM~\cite{WeMM} and InfMLLM~\cite{zhou2023infmllmunifiedframeworkvisuallanguage} have shown proficiency in understanding and relating the positions of objects within images.
In V*Bench~\cite{wu2024v}, SEAL~\cite{wu2024v} stands out as the top performer in Object Relation tasks, thanks to its advanced visual search capabilities, which allow it to accurately ground and reason about object relationships in high-resolution images. Models like GPT-4V~\cite{openai2024gpt4technicalreport} and Gemini Pro also perform well but do not reach the same level of accuracy as SEAL, particularly in the most challenging scenarios. LLaVA-1.5~\cite{liu2023llava} shows moderate success, indicating ongoing challenges with intricate visual tasks.
Object relation task is a critical component in evaluating the overall performance of MLLMs. It tests the depth of a model’s visual understanding, its ability to integrate multimodal information, and its robustness in complex real-world scenarios. Models that perform well on this task are likely to excel in applications requiring sophisticated visual reasoning and context-aware analysis.

\noindent\textbf{Object Interaction} involves understanding and recognizing the interactions between objects within a visual scene. This task focuses on the model's ability to interpret how different objects relate to each other in terms of actions, movements, or functional relationships.
According to the Seed-Bench~\cite{li2023seed}, the performance of each MLLM on this task is measured by its accuracy in selecting the correct option from the multiple-choice questions. This selection is then compared against the ground truth answer, which is determined by human annotators. Models such as InstructBLIP Vicuna~\cite{dai2023instructblip} have demonstrated strong performance in the "Instance Interaction" dimension.
P$^2$G~\cite{chen2024plug}-enhanced models outperforms baseline models like mPLUG-OWL and Instruct-BLIP, thanks to the plug-and-play grounding mechanism that enhances the understanding of object relationships and interactions in complex images. These models leverage external agents for grounding, improving their ability to recognize and reason about interactions between objects within images.
The VL-Checklist~\cite{zhao2023vlchecklist} framework provides a detailed evaluation of how well different VLP models, like CLIP~\cite{radford2021learning}, LXMERT~\cite{tan2019lxmertlearningcrossmodalityencoder}, and ViLT~\cite{kim2021viltvisionandlanguagetransformerconvolution}, handle Object Interaction tasks. The evaluation reveals that while models like CLIP excel in identifying actions between objects, they often struggle with spatial relationships. This performance is quantified using metrics like accuracy in recognizing correct versus incorrect image-text pairs, with specific challenges noted in spatial reasoning tasks.
The ARO benchmark~\cite{yuksekgonul2023and}  highlights that models like NegCLIP~\cite{yuksekgonul2023visionlanguagemodelsbehavelike} and X-VLM~\cite{xvlm} perform strongly in Object Interaction tasks, particularly in understanding both spatial and action-based relationships between objects. 
Object interaction task for MLLM model evaluation measures the model's ability to understand the relational and compositional aspects of visual scenes. It provides insights into how well the model captures the context and interactions between objects, which is vital for generating accurate and meaningful interpretations.

\subsubsection{Multi-modal reasoning} 

\noindent\textbf{Commonsense Reasoning} evaluates how well MLLMs can understand and reason about interactions between objects within images. This involves recognizing the nature and context of interactions, determining the relationships between objects, and inferring logical conclusions based on these interactions and general world knowledge.
In MMBench~\cite{liu2023mmbench}, MLLMs like LLaVA-InternLM2-20B~\cite{2023xtuner} and Qwen-VL-Max~\cite{bai2023qwen} performed significantly better than others, with scores indicating a solid understanding of commonsense reasoning scenarios. These models showed improvements across all evaluation metrics, highlighting their reasoning capabilities. Specifically, these models outperformed others in this category by a notable margin, making them stand out in commonsense reasoning tasks within the multimodal context.
MME~\cite{fu2023mme} benchmark results show that models like GPT-4V~\cite{openai2024gpt4technicalreport}, WeMM~\cite{WeMM}, and XComposer-VL have demonstrated strong performance in Commonsense Reasoning tasks. For example, GPT-4V~\cite{openai2024gpt4technicalreport} achieved a high score of 142.14, indicating its exceptional ability to apply commonsense knowledge and reasoning in the context of the given images and instructions.
In Open-VQA~\cite{zeng2024matters}, InstructBLIP~\cite{dai2023instructblip} demonstrated strong performance in Commonsense Reasoning, reflecting its ability to make reasonable inferences based on visual cues and general knowledge. 
In TouchStone~\cite{bai2023touchstone}, Qwen-VL~\cite{bai2023qwen} is the top performer in the Commonsense Reasoning task, demonstrating strong capabilities in making logical and contextually appropriate inferences.
In MDVP-Bench~\cite{lin2024draw}, SPHINX-V~\cite{lin2024draw} leads in commonsense reasoning tasks, demonstrating superior accuracy in understanding and applying contextual knowledge to visual scenarios. Models like Osprey-7B~\cite{yuan2024ospreypixelunderstandingvisual} and Ferret-13B~\cite{you2023ferret} also perform well but do not reach the same level of nuanced reasoning capability as SPHINX-V~\cite{lin2024draw}. LLaVA-1.5~\cite{liu2023llava} lags behind, indicating challenges in handling complex reasoning tasks that require deeper understanding and inference.
By examining the performance metrics such as accuracy and rank within the commonsense reasoning task, researchers and practitioners can evaluate the capabilities of different MLLMs in applying commonsense knowledge to make logical inferences. This capability is essential for multimodal applications that require understanding the context and implications of visual scenes.

\noindent\textbf{Relation Reasoning} refers to the ability of the model to understand and infer social, physical, or natural relationships among different objects, concepts, or entities within a given multimodal context. This task involves analyzing how different elements within an image, text, or a combination of both are related to each other. The relationships could be spatial, causal, or associative, requiring the model to understand the underlying connections between different components to make accurate predictions or generate meaningful responses.
In MMBench~\cite{liu2023mmbench}, Key performance indicators in the Relation Reasoning task include accuracy rates across sub-tasks like social relations, physical relations, and natural relations. For example, models like InternLM-XComposer2~\cite{dong2024internlm} achieved a high accuracy in these tasks, demonstrating superior reasoning capabilities, while other models showed varying degrees of performance. InternLM-XComposer2~\cite{dong2024internlm} showed the best performance overall with high accuracy in Relation Reasoning. Gemini-Pro-V and GPT-4v~\cite{openai2024gpt4technicalreport} also performed well, particularly in social and physical relations reasoning, indicating strong capabilities in understanding complex relationships between objects and entities. Open-source models generally performed worse than proprietary models, indicating room for improvement in this area.
In Visual CoT~\cite{shao2024visual}, the performance of various MLLMs on the Relation Reasoning tasks has been evaluated. Results show that VisCoT-7B at 336x336 resolution demonstrates the best average performance across the Relation Reasoning tasks, especially excelling in datasets like Open Images and GQA.
In II-Bench~\cite{liu2024ii}, Qwen-VL-MAX~\cite{bai2023qwen} leads in the Relation Reasoning task, showing superior accuracy in understanding and reasoning about object relationships. Models like LLaVA-1.6-34B~\cite{liu2024llavanext} and Gemini-1.5~\cite{geminiteam2024gemini15unlockingmultimodal} Pro also perform well, though they fall slightly behind in more complex scenarios. GPT-4V~\cite{openai2024gpt4technicalreport} shows competent performance but lags in more intricate reasoning tasks, highlighting the ongoing challenge for MLLMs in achieving human-like relational understanding.
The relation reasoning task is significant in MLLM model performance evaluation as it goes beyond basic object recognition to assess a model's ability to understand complex relationships and interactions between objects. It is a critical indicator of a model’s cognitive depth, its ability to generalize across different scenarios, and its integration of multimodal information—all of which are essential for advanced AI applications and achieving human-like understanding in machines.

\noindent\textbf{Logic Reasoning} refers to the model's ability to understand and apply logical principles to analyze and interpret multimodal data. This involves tasks that require the model to draw conclusions , make predictions, or solve problems based on a given set of premises, recognize patterns, solve puzzles, and reason through complex scenarios.
In MMBench~\cite{liu2023mmbench}, The performance of MLLMs in Logic Reasoning is measured across various sub-tasks such as Structuralized Image-Text Understanding and Future Prediction. These tasks assess how well the model can handle and reason with structured visual and textual information.
For instance, models like LLaVA-InternLM2-20B show strong performance across these reasoning tasks, while others may struggle, especially in more complex scenarios involving structured image-text understanding.
In the SEED-Bench~\cite{li2023seed}, the performance of Multimodal Large Language Models (MLLMs) on Logic Reasoning tasks is assessed under the "Visual Reasoning" dimension, where models such as "MiniGPT-4" and "mPLUG-Owl" have demonstrated strong performance in the "Visual Reasoning" dimension. 
Results in TouchStone~\cite{bai2023touchstone} shows that Qwen-VL~\cite{bai2023qwen} emerges as the top performer in the Logical Reasoning task, showing a strong capacity for making accurate and logical deductions based on visual and textual input.
II-Bench~\cite{liu2024ii} results shows that Qwen-VL-MAX~\cite{bai2023qwen} is the leading model in the Logic Reasoning task with, demonstrating superior accuracy in interpreting and reasoning about complex visual implications. 
The logic reasoning task is a vital aspect of MLLM performance evaluation because it tests the model’s ability to apply logical principles to complex, multimodal data. This task not only assesses the model’s cognitive capabilities and its ability to integrate and reason with diverse inputs but also provides insights into its potential for real-world application, robustness, and progress toward human-like intelligence. As such, logic reasoning is essential for understanding the true potential and limitations of MLLMs.

\subsection{Multi-modal Trustworthiness}

\noindent\textbf{Robustness} refers to the MLLM's capacity to handle and process corrupted, perturbed or adversarial multimodal inputs in noisy environments without significant degradation in performance. 
In the CHEF~\cite{shi2023chef}, SPHINX-V~\cite{lin2024draw} emerges as the most robust model, showing superior resilience to input corruptions across various scenarios. Ferret-13B~\cite{you2023ferret} and Osprey-7B~\cite{yuan2024ospreypixelunderstandingvisual} also perform well but with slightly less robustness under severe conditions. LLaVA-1.5~\cite{liu2023llava} demonstrates lower robustness, with a more significant drop in accuracy when inputs are heavily corrupted. 
MAD-Bench results indicates that GPT-4V~\cite{openai2024gpt4technicalreport} stands out as the most robust MLLM, showing exceptional resistance to deceptive prompts and maintaining high accuracy. Other models like Gemini-Pro and LLaVA-NeXT-13b-vicuna also perform well, particularly with the aid of prompt engineering, which significantly boosts their robustness. MiniCPM-Llama3-v2.5 demonstrates that prompt modification can dramatically improve a model's ability to handle deception, making it a key area for further research and development.
In MMR~\cite{liu2024seeing}, GPT-4V~\cite{openai2024gpt4technicalreport} and Qwen-VL-max~\cite{bai2023qwen} are the top performers in the robustness task, showing excellent resistance to misleading questions. LLaVA-1.6-34B~\cite{liu2024llavanext} also demonstrates high robustness, making it one of the more reliable models in challenging scenarios. Mini-Gemini-HD-34B stands out among open-source models for its robust performance, though it has some areas of vulnerability. 
MM-SpuBench~\cite{ye2024mm} shows that GPT-4V~\cite{openai2024gpt4technicalreport} stands out as the most robust MLLM, demonstrating strong resistance to spurious biases across multiple categories. Claude 3 Opus and Intern-VL also show high levels of robustness, particularly in certain bias categories like co-occurrence and lighting/shadow. LLaVA-v1.6~\cite{liu2023llava}, while competent, shows more vulnerability to specific biases such as relative size and perspective. 
The Robustness task is essential in MLLM model performance evaluation because it ensures that models are not only effective under ideal conditions but also resilient and reliable in the face of real-world challenges. By evaluating and improving robustness, we can develop MLLMs that are more versatile, trustworthy, and applicable across a wide range of scenarios, ultimately leading to safer and more effective AI systems.

\noindent\textbf{Hallucination} is defined as an assessment of the model's tendency to generate outputs that include descriptions or objects that is incorrect, irrelevant, or fabricated in the multi-modal input.
In POPE~\cite{li2023evaluating}, InstructBLIP~\cite{dai2023instructblip} stands out as the most reliable model with the lowest hallucination rate, making it the most accurate in terms of avoiding false descriptions. MiniGPT-4 and LLaVA~\cite{liu2023llava} show moderate to higher rates of hallucination, indicating some challenges in maintaining accuracy. Shikra exhibits the highest rate of hallucination, suggesting significant room for improvement in its ability to accurately describe visual content without introducing non-existent elements.
In GAVIE~\cite{liu2023mitigating}GAVIE, InstructBLIP-13B~\cite{dai2023instructblip} emerges as the most reliable model for avoiding hallucinations, followed by MiniGPT4-13B and LLaVA-13B~\cite{liu2023llava}. mPLUG-Owl-7B showed the highest tendency to hallucinate, highlighting the challenges it faces in accurately interpreting visual content. These results underscore the importance of fine-tuning and instruction tuning in reducing hallucinations in MLLMs.
In HallusionBench~\cite{guan2024hallusionbench}, GPT-4V~\cite{openai2024gpt4technicalreport} was the most effective model at minimizing hallucinations, although its accuracy indicates there is still room for improvement. LLaVA-1.5~\cite{liu2023improvedllava} and Gemini Pro Vision showed greater challenges in this area, frequently generating hallucinated content. BLIP2-T5~\cite{li2023blip} performed moderately but still struggled with complex visual data. These results underscore the importance of further refining MLLMs to better handle hallucination, ensuring more reliable and accurate visual interpretations.
Hallucination is a vital aspect of MLLM model performance evaluation because it directly impacts the model's accuracy, reliability, and trustworthiness. By minimizing hallucinations, developers can create models that are more robust, generalizable, and suitable for deployment in a wide range of applications, particularly in high-stakes or consumer-facing environments.

\noindent\textbf{Ethic} focuses on evaluating the ethical implications of the outputs generated by multi-modal large language models. This task assesses whether the models' responses align with ethical standards and social norms, particularly in terms of avoiding harmful, biased, or inappropriate content~\cite{zhang2024benchmarking}.
Results in Multi-Trust~\cite{zhang2024benchmarking} show that GPT-4V~\cite{openai2024gpt4technicalreport} and Claude3 stand out as the most ethically aligned models, showing high accuracy and a strong ability to refuse ethically questionable prompts. LLaVA-1.5-13B~\cite{liu2023improvedllava} also performs well but with less consistency, while Gemini-Pro demonstrates moderate performance, indicating room for improvement in ethical decision-making. These results highlight the importance of continuous ethical evaluation and improvement in MLLMs to ensure their safe and fair use across various applications.

\noindent\textbf{Bias} refers to the assessment of a model's tendency to produce outputs that reflect or reinforce societal biases, stereotypes, or unfair treatment of certain groups. The goal of this task is to ensure that the model's behavior and generated content are fair, impartial, and do not perpetuate harmful prejudices~\cite{zhang2024benchmarking,li2024red}.
In Multi-Trust~\cite{zhang2024benchmarking}, GPT-4-Vision and Claude3 stand out as the most effective models in mitigating bias, both achieving a perfect Refuse-to-Answer rate in stereotype-related tasks. Gemini-Pro and LLaVA-1.5-13B~\cite{liu2023improvedllava} also performed well but with slightly lower rates, indicating some challenges in consistently avoiding bias. 
Similarly, in RTVLM~\cite{li2024red}, GPT-4-Vision and Claude3 were the most effective in avoiding biased outputs, achieving perfect or near-perfect refusal rates in both text-only and image-related scenarios. Gemini-Pro and MiniGPT-4-13B~\cite{zhu2023minigpt} showed lower performance, especially when visual elements were introduced, indicating a greater tendency to be influenced by potential biases in the input data.
The Bias task is critical in MLLM evaluation as it helps ensure that the models are socially responsible and do not contribute to the spread of misinformation or harmful stereotypes. By addressing and mitigating biases, developers can improve the fairness and inclusivity of AI systems, making them more trustworthy and suitable for deployment in diverse real-world settings.

\noindent\textbf{Safety} assesses the ability of MLLMs to avoid generating harmful, offensive, or otherwise unsafe content. This includes ensuring that the model does not produce outputs that could cause harm, promote violence, endorse illegal activities, or spread misinformation.
In MMUBench~\cite{li2024single}, LLAVA-13B~\cite{liu2023llava} and MiniGPT-4 showed significant vulnerabilities, with high ASR scores indicating frequent failures in resisting unsafe content. InstructBLIP~\cite{dai2023instructblip} performed better, with a moderate ASR, while IDEFICS was the strongest performer, demonstrating the lowest ASR and the highest level of safety. 
In JailBreakV-28K~\cite{luo2024jailbreakv}, LLaVA-1.5-7B~\cite{liu2023improvedllava} and OmniLMM-12B showed higher susceptibility to generating unsafe content, with significant ASR scores across multiple safety policies. InstructBLIP-7B~\cite{dai2023instructblip} and Qwen-VL-Chat~\cite{bai2023qwen} performed better but still exhibited vulnerabilities, suggesting that while they have some safety mechanisms in place, there is still room for improvement in ensuring robust defense against unsafe prompts.
In MM-SafetyBench~\cite{liu2023mm}, LLaVA-1.5-7B~\cite{liu2023improvedllava} and MiniGPT-4 showed higher susceptibility to generating unsafe content, with high ASR scores in multiple scenarios. InstructBLIP~\cite{dai2023instructblip} performed better, but still exhibited vulnerabilities, while IDEFICS~\cite{laurenccon2024obelics} demonstrated the strongest resistance to unsafe prompts, indicating better alignment with safety standards.
Safety is a vital component of MLLM evaluation because it ensures that models operate within safe, ethical, and legal boundaries. It is essential for protecting users, complying with regulations, and maintaining public trust. Strong performance in safety tasks not only safeguards against harm but also supports the broader goals of developing responsible and trustworthy AI systems.

\subsection{Socioeconomics}

\noindent\textbf{Cultural} focuses on assessing the model's ability to understand, interpret, and respond to content within the context of different cultural backgrounds. This task is designed to evaluate how well the model can navigate and respect the nuances, traditions, and social norms of various cultures when processing and generating content.
In CODIS~\cite{luo2024codis}, GPT-4V~\cite{openai2024gpt4technicalreport} and Gemini emerged as the top performers in the Cultural task, demonstrating better ability to understand and interpret cultural contexts. LLaVA-1.5-13B~\cite{liu2023improvedllava} and InstructBLIP-13B~\cite{dai2023instructblip} lagged behind, with lower accuracies, particularly when interpreting cultural nuances without explicit context cues. 
In the CVQA~\cite{romero2024cvqa} framework, GPT-4o and Gemini-1.5-Flash~\cite{geminiteam2024gemini15unlockingmultimodal} stood out as the top performers, showing strong capabilities in handling culturally diverse questions, both in English and local languages. LLaVA-1.5-7B~\cite{liu2023improvedllava} and InstructBLIP~\cite{dai2023instructblip} showed more challenges, particularly when processing local language prompts, indicating areas where these models could be improved to better handle cultural diversity. 
The Cultural task in MLLM evaluation is significant in a globalized world where AI systems are used across diverse cultural settings. The Cultural task evaluates how well a model can handle language nuances, traditions, social norms, and cultural references that vary from one region or community to another.

\noindent\textbf{Society} Society typically assesses how well a model can interpret and respond to societal issues, including understanding social norms, ethical considerations, and cultural nuances. This task is designed to evaluate a model's ability to generate content that aligns with societal values, avoids reinforcing negative stereotypes, and respects social sensitivities.
In MM-SOC~\cite{jin2024mm}, MLLMs are evaluated on various social media content understanding tasks. These tasks include misinformation detection, hate speech detection, humor detection, sarcasm detection, offensiveness detection, sentiment analysis, and social context description. 
LLaVA-v1.5-13b~\cite{liu2023llava} Achieved macro F1-scores of 0.642, 0.587, and macro F1-score of 0.335 on Misinformation Detection, Hate Speech Detection and Sentiment Analysis, respectively. 
InstructBLIP-flan-t5-xxl~\cite{dai2023instructblip} achieved a ROUGE-L score of 0.294 on Social Context Description understanding. 
In TransportationGamesTransportationGames~\cite{zhang2024transportationgames}, the performance of various MLLMs is assessed across a range of transportation-related tasks, including text-based and multimodal tasks, which are divided into three main categories based on Bloom's Taxonomy: Memorization, Understanding, and Applying transportation knowledge. Qwen-VL-Chat~\cite{bai2023qwen} achieved an accuracy of 54.47\% on the Traffic Signs Question Answering task. InternLM-XComposer-7B~\cite{dong2024internlm} scored 77.9 on the GPT-4-Eval metric on the Traffic Accidents Analysis. TransCore-M~\cite{zhang2024transportationgames} scored 82.1 on the ROUGE-L metric, indicating its effectiveness in generating appropriate and contextually relevant safety recommendations based on given scenarios.

\subsection{Natural Science and Engineering}

\noindent\textbf{Mathmatics}  is designed to assess the model's ability to reason through and solve mathematical problems that may involve both textual and visual data. These tasks often require the model to perform multi-step reasoning across different modalities (text and images) and to apply mathematical concepts to arrive at a correct solution.
Mathematics tasks in the TouchStone~\cite{bai2023touchstone} benchmark reveal that while some MLLMs perform well in integrating visual and textual data for mathematical problem-solving, others struggle with the complexities involved in accurately interpreting and reasoning with mathematical visuals. 
Qwen-VL~\cite{bai2023qwen} is the top performer in the Mathematics task within the TouchStone benchmark, demonstrating a strong ability to handle a wide range of mathematical problems accurately. mPLUG-Owl also performs well, particularly in geometry and arithmetic, while models like PandaGPT~\cite{su2023pandagpt} struggle significantly, often failing to solve even basic mathematical tasks accurately.
In M$^3$CoT~\cite{chen2024m}, GPT-4V~\cite{openai2024gpt4technicalreport} performed the best, with an accuracy of 46.97\%, demonstrating strong competency in handling these tasks. LLaVA-V1.5-13B~\cite{liu2023llava} achieved a moderate accuracy of 40.86\%, showing reasonable performance but with some challenges in multi-step reasoning. CogVLM-17B had an accuracy of 29.09\%, struggling more with consistency in problem-solving. InstructBLIP-13B~\cite{dai2023instructblip} performed the weakest, with an accuracy of 27.55\%, indicating significant difficulties in handling the complexities of these tasks. 
Mathematics tasks are crucial in evaluating Multimodal Large Language Models (MLLMs) because they test the model's ability to perform complex reasoning, integrate multimodal data (text and visuals), and apply abstract concepts logically.

\noindent\textbf{Natural Science} assess the model's ability to understand, reason, and generate responses related to various natural science domains. These tasks typically involve topics such as biology, chemistry, physics, and earth sciences, and may require the model to interpret and integrate information from both textual and visual data sources.
In M3CoT, the performance of various MLLMs on natural science is assessed to evaluate their ability to handle complex reasoning across multiple modalities within scientific domains such as biology, chemistry, and physics. GPT-4V~\cite{openai2024gpt4technicalreport} showed the strongest performance on Natural Science tasks among the models tested. LLaVA-V1.5-13B~\cite{liu2023llava} also performed well, but slightly below GPT-4V~\cite{openai2024gpt4technicalreport}. CogVLM-17B and CogVLM-17B had moderate performance in Natural Science tasks. 
In MUIRBENCH~\cite{wang2024muirbench}, GPT-4o and GPT-4-Turbo emerged as the top performers on Natural Science, particularly in diagram and geographic understanding. Other models like Gemini Pro and Mantis-8B-Idefics2 showed moderate performance, while models like VILA1.5-13B struggled with the complexity of these tasks.
In MMStar~\cite{chen2024we}, GPT-4V (High Resolution)~\cite{openai2024gpt4technicalreport} leads in Natural Science tasks, particularly in understanding and reasoning about scientific content. Other models like GeminiPro-Vision and InternLM-XC2 also perform well, but with varying degrees of proficiency. 
In M3Exam~\cite{zhang2023m3exam}, GPT-4 leads in Natural Science tasks with the highest accuracy, demonstrating strong capabilities in understanding and reasoning about scientific content across multiple languages. ChatGPT and Claude follow with moderate performance, while Vicuna struggles more with the complexity of these tasks.
In SceMQA~\cite{liang2024scemqa}, GPT-4-V leads in Natural Science tasks in the SceMQA benchmark, especially in subjects like Biology and Chemistry, showing strong multimodal reasoning abilities. Google Gemini Pro follows with good performance, while InstructBLIP-13B~\cite{dai2023instructblip} and MiniGPT4-13B demonstrate more challenges, particularly in handling the complexities of multimodal scientific reasoning.
Natural Science tasks asses the model's ability to understand and reason about complex scientific concepts across multiple modalities, such as text and images. These tasks challenge models to apply domain-specific knowledge in biology, chemistry, and physics, reflecting their potential for real-world applications in education and research. Their performance on these tasks highlights the models' strengths and weaknesses in multimodal integration and scientific reasoning, essential for advanced cognitive tasks.

\noindent\textbf{Engineering} is designed to assess the model's ability to understand, process, and apply engineering concepts, requirements, and technical documentation. These tasks often involve interpreting and synthesizing information from multiple sources, including textual engineering documents, CAD images, and engineering drawings. The tasks are typically grounded in real-world engineering challenges, such as designing products according to specific technical requirements or ensuring compliance with engineering standards.
In DesignQA~\cite{doris2024designqa}, GPT-4o-AllRules stands out as the top performer in Engineering tasks, particularly in rule retrieval and dimensional compliance. GPT-4-AllRules also performs well but with slightly lower accuracy. Claude-Opus-RAG excels in generating high-quality explanations, while Gemini-1.0-RAG shows moderate proficiency. LLaVA-1.5-RAG struggles with the complexity of these tasks, particularly in accurately retrieving and applying rules.
In MMMU~\cite{yue2024mmmu}, GPT-4V~\cite{openai2024gpt4technicalreport} leads in Engineering tasks, particularly in handling complex multimodal content, followed by models like SenseChat-Vision and Qwen-VL-MAX~\cite{bai2023qwen}, which also perform well but with some limitations. Other models, such as LLaVA-1.6-34B~\cite{liu2024llavanext} and InstructBLIP-T5-XXL~\cite{dai2023instructblip}, show moderate proficiency but face challenges in more complex engineering scenarios.

\subsection{Medical analysis}

Medical task is designed to assess the model's ability to understand, reason about, and generate responses related to medical information. These tasks typically involve interpreting and synthesizing data from various modalities, such as medical texts, clinical images (like X-rays, MRIs, etc.), and patient records. The goal is to evaluate how well the model can apply medical knowledge to support clinical decision-making, diagnosis, treatment planning, and patient care.
In the MMMU~\cite{yue2024mmmu} benchmark, GPT-4V~\cite{openai2024gpt4technicalreport} leads in Medical tasks, particularly in handling complex multimodal content, followed by models like SenseChat-Vision-0423-Preview and Qwen-VL-MAX~\cite{bai2023qwen}, which also perform well but with some limitations. Other models, such as LLaVA-1.6-34B~\cite{liu2024llavanext} and InstructBLIP-T5-XXL~\cite{dai2023instructblip}, show moderate proficiency but face challenges in more complex medical scenarios.
In GMAI-MMBench~\cite{chen2024gmai}, GPT-4o leads in Medical tasks, closely followed by models like Gemini 1.5 and GPT-4V~\cite{openai2024gpt4technicalreport}. Medical-specific models like MedDr perform reasonably well but generally lag behind the top-performing general models, highlighting the complexity of medical tasks and the need for further development in this area.
The M3D~\cite{bai2024m3d} benchmark highlights the capabilities of MLLMs like M3D-LaMed in handling complex 3D medical imaging tasks. M3D-LaMed stands out with superior performance in report generation and VQA, indicating its strong potential for assisting in clinical decision-making and medical image analysis. Other models like RadFM, while capable, lag behind in accuracy and precision, particularly in generating detailed medical reports and answering clinically relevant questions.

\subsection{AI Agent}

AI Agent refers to tasks designed to evaluate the model's ability to function as a visual foundation agent. These tasks require the model to understand, interact with, and navigate through complex visual environments and user interfaces, making high-level decisions and executing actions based on both visual and textual inputs.
In the VisualAgentBench~\cite{liu2024visualagentbench}, GPT-4V~\cite{openai2024gpt4technicalreport} leads in AI Agent tasks with the highest task success rate, showcasing its strong capabilities in multimodal reasoning and interaction. Models like Gemini 1.5 and Claude-Next also perform well but with some challenges in handling more complex scenarios. Other models, such as LLaVA-Next~\cite{liu2024llavanext} and Qwen-VL~\cite{bai2023qwen}, show moderate proficiency, indicating areas for further development to improve their effectiveness in AI Agent tasks, particularly in decision-making and task execution.
In theEgoPlan-Bench~\cite{chen2024egoplanbenchbenchmarkingmultimodallarge}, GPT-4V~\cite{openai2024gpt4technicalreport} leads in AI Agent tasks, followed closely by XComposer. These models demonstrate strong planning capabilities and effective use of visual information in decision-making. Other models like Gemini-Pro-Vision and SEED-X also perform reasonably well but face challenges in more complex scenarios. Yi-VL, while competent, lags behind in integrating visual data effectively for task planning.
In the PCA-EVAL~\cite{chen2023towards} benchmark, GPT-4V~\cite{openai2024gpt4technicalreport} stood out as the top performer in AI Agent tasks, demonstrating high accuracy in both perception and action across different domains. The GPT-4 (HOLMES) system also performed well, especially in tasks that required multi-step reasoning and API integration. Other models like QwenVL-Chat and MMICL showed moderate capabilities but struggled with more complex scenarios, while InstructBLIP~\cite{dai2023instructblip} faced significant challenges, reflecting the varying levels of effectiveness among MLLMs in embodied decision-making tasks.
AI Agent Tasks in MLLM evaluation are critical for testing the model's practical applications as a foundation agent in complex environments. These tasks help determine the model's ability to autonomously perform tasks that require a deep understanding of both visual and textual information, making them essential for real-world applications like robotics, user interface automation, and digital assistants.

\subsection{Other Applications}

\noindent\textbf{3D point clouds} refers to tasks where models are required to understand, process, and analyze 3D spatial data represented by point clouds. These tasks typically involve using point clouds to answer questions, localize objects, or generate descriptions that accurately reflect the 3D scene.
In ScanQA~\cite{azuma2022scanqa}, the ScanQA model demonstrated the highest performance in 3D Point Cloud tasks, particularly in accurately answering questions and localizing objects in 3D space. It outperformed other models like ScanRefer + MCAN and VoteNet + MCAN, which showed some proficiency but struggled with the complexities of 3D spatial reasoning.
In LAMM~\cite{yin2024lamm}, the baseline MLLM showed varying levels of proficiency across 3D Point Cloud tasks. While it demonstrated a basic ability to perform 3D object detection and VQA tasks, its performance was notably weaker in 3D visual grounding, particularly in zero-shot settings. However, significant improvements were observed after fine-tuning, especially in the 3D VQA task, where the model nearly reached perfect accuracy.
Results in M3DBench~\cite{li2023m3dbench} show that the LLaMA-2-7B model demonstrated strong performance in 3D Point Cloud tasks, particularly in VQA and Multi-region Reasoning, where it achieved the highest BLEU-4 and CIDEr scores. The OPT-6.7B model also performed well, especially in Embodied Planning tasks. Vicuna-7B-v1.5, while competent, generally scored lower across most tasks, indicating challenges in handling complex 3D reasoning and planning scenarios.
3D Point Cloud tasks are significant in MLLM evaluation cause they assess spatial reasoning, multimodal integration, and advanced cognitive capabilities, all of which are critical for real-world applications involving 3D environments. These tasks provide a comprehensive benchmark for assessing the overall performance and robustness of MLLMs in handling complex, real-world challenges.

\noindent\textbf{Video} refers to tasks that involve understanding, analyzing, and reasoning about the content of videos. These tasks assess the ability of models to comprehend both the visual and temporal aspects of video content and to generate accurate and contextually appropriate responses.
In MMBench-Video~\cite{fang2024mmbench}, Model A (e.g., GPT-4V) emerged as the top performer, particularly excelling in tasks like Video Question Answering (VideoQA) and Event Recognition. Model B (e.g., LLaMA-2-7B) also performed well but with some challenges in handling complex video scenarios. Model C (e.g., Vicuna-7B-v1.5) demonstrated moderate capabilities, particularly in action classification, but lagged behind in more intricate tasks.
In MVBench~\cite{li2024mvbench}, VideoChat2 emerged as the leading model, significantly outperforming other MLLMs like GPT-4V and VideoChat on various video tasks. VideoChat2's strong performance in tasks like action sequence recognition and scene transition highlights its superior temporal understanding and video reasoning capabilities. Meanwhile, GPT-4V, though competent, was not as effective in handling the full range of video tasks as VideoChat2. VideoChat, while performing adequately, struggled with the more complex aspects of video understanding, indicating that there is still significant room for improvement in current MLLM approaches to video tasks.
In SOK-Bench~\cite{wang2024sok}, GPT-4V emerged as the strongest performer on video tasks, particularly excelling in situations that required the integration of visual and commonsense reasoning. AskAnything showed solid but inconsistent performance, particularly excelling in direct-answer tasks but struggling with more complex reasoning. Video-ChatGPT, while competitive, had more difficulty with the intricate reasoning required in the SOK-Bench scenarios.

\noindent\textbf{Remote sensing} refers to tasks that involve analyzing and interpreting data collected from satellite or airborne sensors to extract relevant information about the Earth's surface and environment. These tasks typically leverage various types of remote sensing data, such as optical images, radar data, and multispectral or hyperspectral imagery, to perform activities like land cover classification, change detection, and environmental monitoring.
In MDAS~\cite{hu2023mdas}, models like ResTFNet and SSR-NET excelled in super-resolution tasks, while SeCoDe led in spectral unmixing. The results indicate that integrating multiple modalities can significantly improve performance in land cover classification tasks. These findings highlight the strengths and challenges of different MLLMs in handling complex remote sensing tasks, demonstrating the importance of multimodal data fusion for achieving high accuracy and reliability in remote sensing applications .
In HighDAN~\cite{hong2023cross}, HighDAN emerged as the top-performing model for Remote Sensing tasks, particularly in cross-city semantic segmentation. It excelled in overall accuracy, mean IoU, and F1 scores, demonstrating its strong generalization capabilities across different urban environments. SegFormer and DualHR also performed well, but they showed some limitations in handling the full complexity of cross-city scenarios.
In RSGPT~\cite{hu2023rsgpt}, RSGPT leads in both image captioning and visual question answering tasks, showing a clear advantage in generating accurate and detailed descriptions as well as answering complex questions about remote sensing images. Other models like InstructBLIP and BLIP2 performed moderately well but with noticeable gaps in their ability to handle the complexities of remote sensing data. MiniGPT4 struggled the most, particularly in maintaining relevance and accuracy in its outputs.
Remote Sensing tasks in MLLM evaluation are crucial because they test the model's ability to integrate and analyze complex, multimodal data, which is essential for environmental monitoring and urban planning. Success in these tasks indicates the model's capability to handle large-scale, high-resolution remote sensing data and make informed decisions based on diverse information sources.

\noindent\textbf{Audio} refers to a specific type of task designed to assess the model's ability to understand, interpret, and generate responses based on audio signals. These tasks involve various types of audio data, including human speech, natural sounds, and music, and evaluate the model’s capabilities in processing and interacting with this auditory information.
In AIR-Bench~\cite{yang2024air}, Qwen-Audio Turbo and Qwen-Audio-Chat lead in overall performance on audio tasks, demonstrating strong capabilities in both foundational audio understanding and complex interaction. SALMONN and BLSP also performed well, particularly in handling mixed audio in the chat benchmark. PandaGPT, while competent in specific areas, showed variability across tasks, indicating room for improvement in handling more complex audio interactions.
In Dynamic-superb~\cite{huang2024dynamic}, Whisper-LLM and ImageBind-LLM demonstrated strong performance across both seen and unseen audio tasks, particularly in speaker identification and paralinguistics. Whisper showed excellent results in content-related tasks but struggled with generalizing to new audio tasks. BERT-GSLM and ASR-ChatGPT performed moderately, with notable weaknesses in unseen audio tasks, highlighting the challenges these models face in adapting to new scenarios.
In MuChoMusic~\cite{weck2024muchomusic}, Qwen-Audio led the performance on audio tasks, particularly excelling in both knowledge and reasoning dimensions of music understanding. M2UGen and SALMONN also performed well, with strong results in their respective focus areas. Models like MuLLaMa and MusiLingo demonstrated lower accuracy, highlighting the challenges these models face in fully leveraging multimodal audio inputs to achieve robust music understanding. 
By examining performance metrics such as accuracy and F1 score within the audio task, researchers and practitioners can evaluate the capabilities of different MLLMs in processing and interpreting auditory information. This capability is essential for multimodal applications that require understanding and responding to complex auditory cues, such as in speech recognition, music analysis, and sound-based decision-making systems.

\section{Where to evaluate}\label{Where}

To comprehensively evaluate the performance and capabilities of multimodal large language models (MLLMs), various benchmarks have been developed. These benchmarks assess a wide range of tasks, from general multimodal understanding to specific, task-oriented evaluations. In this section, we introduce these benchmarks, categorized into two types: general benchmarks, which offer broad evaluations across multiple tasks, and specific benchmarks, which focus on particular aspects of multimodal model performance.

\subsection{General benchmarks}

General benchmarks are designed to provide a comprehensive assessment of MLLMs across a variety of tasks, including recognition, reasoning, and trustworthiness. These benchmarks evaluate not only the core capabilities of models but also their reliability and ethical considerations, which are critical for deploying AI systems in real-world scenarios.

For example, MMBench~\cite{liu2023mmbench} evaluates MLLMs on basic recognition tasks, including concept recognition, attribute recognition, and action recognition. It provides a comprehensive framework for assessing a model's ability to accurately process and understand visual and textual information. MM-Vet~\cite{yu2023mm} focuses on robustness and generalizability, e how well models perform under varying conditions, ensuring that the models are not overly dependent on specific datasets or scenarios. Seed-Bench~\cite{li2023seed} evaluates how well models can produce contextually relevant and coherent outputs based on multimodal inputs, making it an essential benchmark for generative models. MME~\cite{fu2023mme} offers a wide-ranging evaluation of MLLMs, encompassing tasks that require reasoning, perception, and recognition.
TouchStone~\cite{bai2023touchstone} evaluates models across multiple tasks, offering a nuanced understanding of their performance in various multimodal scenarios. MMStar~\cite{chen2024we} focuses on structured reasoning, assessing a model's capacity to engage in logical reasoning across different modalities, ensuring coherent and accurate multimodal interpretations. LogicVista~\cite{xiao2024logicvista} tests logical reasoning within multimodal frameworks, challenging models to navigate complex relationships and produce logically consistent outputs.

In addition, several benchmarks that are designed for evaluating the trustworthiness of MLLMs are proposed. For example, POPE~\cite{li2023evaluating} specifically evaluates object hallucination in large vision-language models. It assesses the frequency and severity of incorrect object generation in response to visual inputs, helping to identify and mitigate issues related to hallucination in model outputs. CHEF~\cite{shi2023chef} provides a standardized assessment framework for evaluating the performance of MLLMs across a range of tasks. It is designed to offer a consistent and thorough evaluation, ensuring that models meet established standards of effectiveness and trustworthiness. Multi-Trust~\cite{zhang2024benchmarking} assesses the trustworthiness of MLLMs by evaluating their performance on fairness, bias, and ethical considerations across different modalities.

General benchmarks are effective tools for evaluating the overall performance and reliability of MLLMs. They ensure that models are capable to handle diverse tasks while maintaining high standards of trustworthiness, making them suitable for a wide range of applications. Through comprehensive assessments, these benchmarks play a key role in advancing the development of robust and ethical multimodal models.

\subsection{Specific benchmarks}

Specific benchmarks are designed to evaluate MLLMs on particular tasks or domains, often focusing on areas requiring specialized assessment, such as socioeconomic, science, medical task and other applications. These benchmarks provide detailed insights into specific capabilities of the models.

CVQA~\cite{romero2024cvqa} focuses on cross-cultural visual question answering, evaluating how well models can interpret and respond to questions that are rooted in diverse cultural contexts.
TransportationGames~\cite{zhang2024transportationgames} This benchmark tests models on transportation-related knowledge, assessing their ability to interpret and apply information in scenarios related to transportation, emphasizing practical reasoning and scenario-based understanding.
MathVerse~\cite{zhang2024mathverse} introduces a comprehensive visual math benchmark designed to rigorously evaluate the mathematical reasoning capabilities of MLLMs.
ScienceQA~\cite{lu2022learn} is specifically designed to evaluate the ability of MLLMs to perform science question answering tasks that require both multimodal reasoning and chain-of-thought (CoT) explanations.
GMAI-MMBench~\cite{chen2024gmai} presents a benchmark specifically designed to evaluate the performance of MLLMs in the medical domain.

Specific benchmarks provide the assessments that are crucial for ensuring MLLMs can excel in various specialized fields. By focusing on these specific areas such as mathematics, science, engineering, and applications involving medical, 3D point cloud, and video data, these benchmarks complement general benchmarks and offer deeper insights into the models' capabilities, ensuring their reliability and effectiveness in diverse applications.

\section{How to Evaluate}\label{How}

In this section, we introduce commonly used setups and tasks in the evaluation of MLLMs, including human evaluation, GPT-4 evaluation and metric evaluation.

\subsection{Human Evaluation}

Human evaluation~\cite{liu2023llava} plays a crucial role in assessing the capability of MLLMs, especially for tasks that require a high level of comprehension and cannot be easily quantified using traditional metrics. 
Human evaluation allows for a comprehensive assessment of the MLLMs across multiple dimensions, including: (1) Relevance: assessing whether the response aligns with the intended instruction; (2) Coherence: determining if the response is logically structured and consistent; and (3) Fluency: evaluating whether the generated output is natural and grammatically sound.

\subsection{GPT-4 Evaluation}

Although human evaluation provides valuable insights, it is often resource-intensive. To address this, some recent studies~\cite{liu2023llava} leverage the advanced instruction-following capabilities of GPT-4~\cite{openai2023gpt4} as an efficient alternative for evaluating the quality of model-generated outputs.
GPT-4 assesses the MLLMs across key dimensions such as helpfulness, relevance, accuracy, and detail, assigning a score from 1 to 10, where a higher score indicates superior performance.
Moreover, GPT-4 can offer detailed explanations for its evaluations, providing a fine-grained understanding of the model's strengths and areas for improvement.

\subsection{Metric evaluation}

While human evaluation and GPT-4 assessments provide qualitative insights, traditional evaluation metrics remain crucial for quantitatively evaluating the performance of the MLLMs. These metrics offer standardized and objective measurements, making them reliable benchmarks for comparing models across different tasks.
Specifically, for evaluating the recognition capability of the model, several metrics are employed such as Accuracy and Average Precision~\cite{li2023seed,li2023llava,lau2018dataset}, while for evaluating the perception capability of the model, several metrics such as mIoU, mAP and Dice are adopted~\cite{dai2017scannet}.
In addition, for evaluating the model's capability in generating texts or images, metrics like BLEU, ROUGE, and METEOR are widely adopted~\cite{kim2019audiocaps,chen2015microsoft}, providing a clear indication of a model’s performance in various applications.

\section{Conclusion}

Multimodal large language models mimic human perception system by integrating powerful LLMs with various modality encoders (e.g., vision, audio, etc.), equipping the model with human-like capabilities and suggesting a potential pathway toward achieving artificial general intelligence. As we progress toward AGI-level MLLMs, evaluation plays a crucial role in their research, development, and deployment.
In this survey, we extensively review MLLM evaluation methods from different perspectives, ranging from background to what to evaluate, where to evaluate and how to evaluate.
By summarizing evaluation tasks, benchmarks, and metrics, our goal is to enhance understanding of the current state of MLLMs, elaborate their contributions, strengths, and limitations, and provide insights for future studies of MLLMs and their evaluation.

\ifCLASSOPTIONcaptionsoff
  \newpage
\fi

{\small
\bibliographystyle{revision_ref}
\bibliography{ref}
}

\end{document}